%% file: samplepaper.tex
\documentclass[runningheads]{llncs}
\usepackage[T1]{fontenc}
\usepackage{graphicx}
\usepackage{amsmath,amssymb}
\usepackage{wrapfig}
\usepackage{booktabs}
\usepackage{multirow}
\usepackage{algorithm}
\usepackage{algpseudocode}
\usepackage{needspace}
\usepackage{hyperref}
\usepackage{pgf}
\usepackage{tabularx}
\usepackage{makecell}
\usepackage{siunitx,colortbl}
\usepackage{setspace}
\usepackage{adjustbox}

\newcommand{\method}{Rapid-Reflex Async-Reflect Agent}
\newcommand{\met}{RRARA}
\begin{document}
\title{Reflex First, Reflect Later: Latency-Aware Embodied LLM Agents for Dynamic Response}
\titlerunning{RRARA: Latency-Aware Embodied LLM Agents for Dynamic Response}
%

\author{
Yangqing Zheng \and
Shunqi Mao \and
Dingxin Zhang \and
Weidong Cai
}

\authorrunning{Y. Zheng et al.}

\institute{The University of Sydney}

\maketitle              
\begin{abstract}
Large language models (LLMs) have substantially improved the planning capabilities of embodied agents, enabling their deployment in dynamic and safety-critical environments. However, these settings expose a critical limitation: inference latency. Delayed LLM responses can weaken real-time responsiveness and misalign agent reasoning with rapidly changing environmental states. This paper systematically studies the impact of inference latency on LLM-based embodied agents in dynamic environments. We introduce an FPS-based Time Conversion Mechanism (TCM) that maps inference time to elapsed simulation time, allowing computational delays to directly affect environmental evolution and agent outcomes. We instantiate this protocol in HAZARD and introduce Response Latency (RL) and Latency-to-Action Ratio (LAR) to evaluate agent responsiveness. Building on this framework, we propose the Rapid-Reflex Async-Reflect Agent (RRARA), which integrates rapid reflexive actions with asynchronous LLM reflection to mitigate latency-induced errors. We further introduce an LLM-based PrePlanner that generates cached object-centric subgoals, reducing repeated LLM calls while retaining the model's high-level reasoning capability. Experiments show that accounting for inference latency substantially changes embodied-agent performance and that RRARA achieves a stronger balance between decision quality and responsiveness.

\end{abstract}
\section{Introduction}

Recent LLMs~\cite{gpt4,claude3,gemini2023} have emerged as powerful high-level planners for embodied agents. Unlike static reasoning tasks, embodied agents must act under real-time constraints while their environments continue to evolve. This is especially important in highly dynamic settings, such as hazard scenarios, multi-agent coordination, and search-and-rescue, where object states, traversability, and task rewards may change during deliberation.

Existing LLM-based embodied agents often improve decision quality by using larger models, longer prompts, chain-of-thought reasoning, or test-time scaling. These strategies improve semantic reasoning but also increase inference latency. In dynamic environments, latency creates a time gap between the state observed by the agent and the state in which the selected action is eventually executed. As a result, a seemingly better plan may become stale before it can be acted upon.

\begin{figure}
\centering
\includegraphics[width=0.8\textwidth]{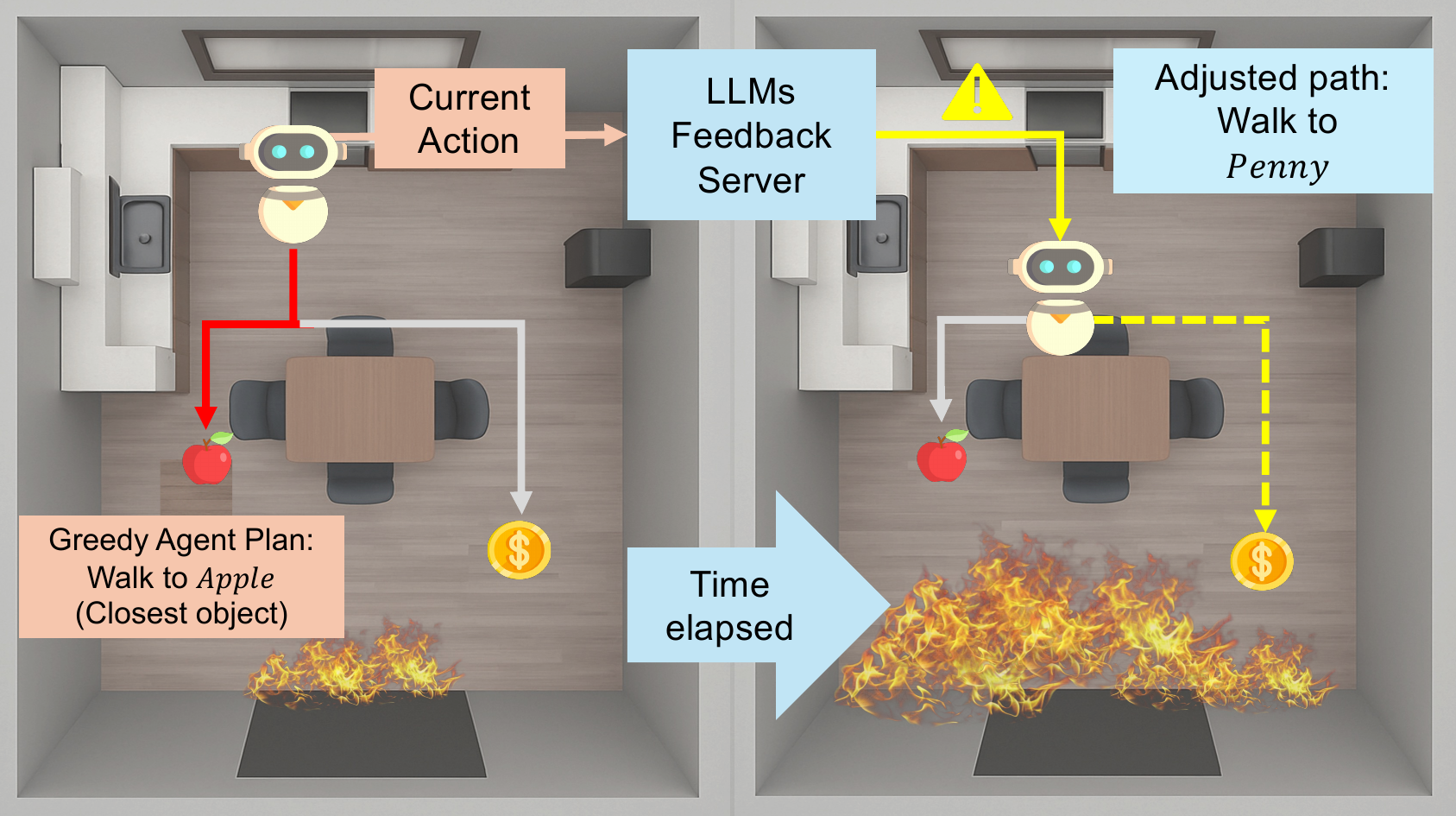}
\caption{\met{} reflective stage. The rapid reflex agent executes an immediate move while the reflective LLM refines this into a goal-aligned plan.} 
\label{fig:teaser}
\end{figure}

Many current benchmarks and evaluation protocols~\cite{alfred,zhou2024hazard,embodiedbenchmark} do not fully capture this effect. Many simulators effectively freeze the environment during model inference or report task success without accounting for wall-clock decision time. We address this gap with a latency-aware evaluation protocol based on a Time Conversion Mechanism, which maps inference delay into simulator frame advancement. This allows us to measure how response latency changes embodied behavior and task outcomes.


Motivated by this observation, we propose the Rapid-Reflex Async-Reflect Agent. Instead of waiting for every LLM decision before acting, RRARA immediately executes low-latency reflexive actions and uses asynchronous LLM reflection to validate, revise, or redirect ongoing behavior. We further introduce an object-centric LLM-PrePlanner that caches future subgoals, reducing repeated LLM calls while preserving long-horizon reasoning.

Our main contributions are:
\renewcommand{\labelitemi}{\textbullet}
\begin{itemize}
    \item We identify and quantify inference latency as a critical bottleneck for LLM-based embodied agents in highly dynamic environments.

    \item We introduce a latency-aware evaluation protocol comprising the Time Conversion Mechanism (TCM), Response Latency (RL), and the Latency-to-Action Ratio (LAR).

    \item We propose RRARA, a reflex-first asynchronous architecture that enables immediate action while using delayed LLM reasoning for plan validation and correction. Under TCM, RRARA achieves an average value rate of 0.44, compared with 0.25 for the strongest LLM baseline.
    \item We empirically evaluate multiple LLM and non-LLM agents on HAZARD and show that RRARA improves average latency-aware performance across the evaluated fire, flood, and wind scenarios.
\end{itemize}

\section{Related Work}
Recent advances in large language models (LLMs) and multimodal large language models (MLLMs) have substantially expanded the scope of embodied artificial intelligence~\cite{gpt4,claude3,llama,rt-1}. 
Beyond serving as text generators, language models are increasingly used as high-level planners, decision makers, and semantic reasoners for embodied agents. 
By leveraging commonsense knowledge and instruction-following ability, LLM-based agents can decompose natural language goals, reason over task constraints, select high-level actions, and interface with perception and control modules.

A growing line of work studies how LLMs can be integrated into embodied control pipelines. Training-free methods such as TANGO compose modular embodied primitives for open-world tasks~\cite{ziliotto2025tango}, while hierarchical approaches such as EPO optimize embodied agents through environment feedback and preference signals~\cite{zhao2024epo}. Other systems explore dual-process or hierarchical architectures that connect deliberative reasoning with low-level execution~\cite{huang2025thinkact}. These methods demonstrate the potential of LLM-based reasoning for embodied control. However, their primary focus is improving task completion or policy quality, while the temporal cost of LLM inference is often treated as an implementation detail rather than a central factor affecting agent behavior.

Widely used benchmarks for LLM- and MLLM-based embodied agents evaluate visual perception, spatial reasoning, long-horizon planning, and embodied execution~\cite{yang2025embodiedbench,alfred}. However, many benchmarks remain latency-unaware, treating the environment as static while the agent reasons.
This assumption limits their ability to represent dynamic embodied settings in which hazards and object states may continue evolving during inference.
HAZARD evaluates agents in dynamic fire, flood, and wind scenarios~\cite{zhou2024hazard}, while Robotouille studies asynchronous planning with action delays, interruptions, and long-horizon dependencies~\cite{2025robotouille}.
These works demonstrate that temporal constraints expose agent failures beyond final task completion, but they do not directly quantify how model inference latency affects embodied decision-making.
Our work explicitly studies latency-aware embodied decision-making in dynamic hazard environments. We introduce a frame-based time conversion mechanism that maps model inference delay to environment time, allowing latency-induced state drift to be measured during evaluation. Building on this protocol, we propose a hybrid agent architecture that combines rapid reflexive responses with asynchronous reflective planning, enabling the agent to act immediately while incorporating delayed LLM reasoning only when it remains temporally valid.

\section{Method}

\begin{figure}[h]
\includegraphics[width=0.9\linewidth]{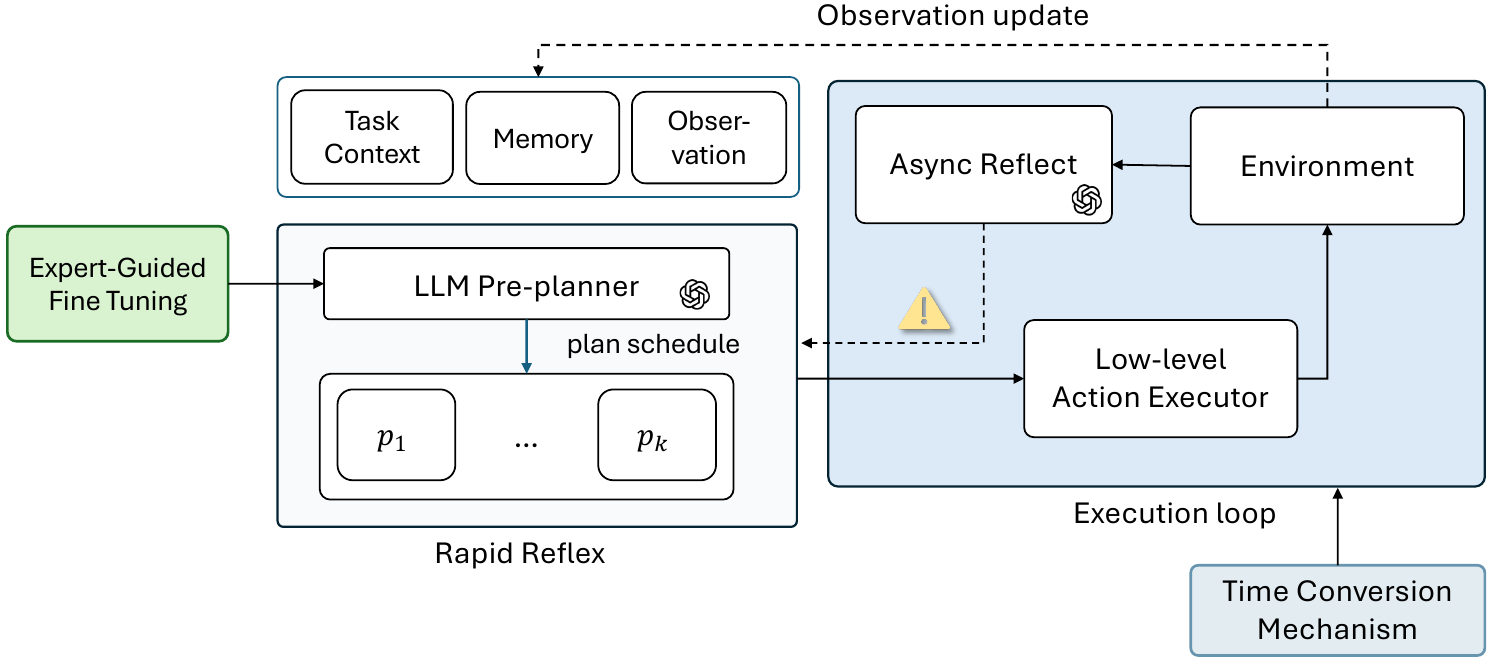}
\caption{Overview of the proposed \met{} framework.  The rapid-reflex module executes a cached plan, while the asynchronous reflector monitors environmental changes and provides corrective feedback without interrupting execution.}
\label{fig:framework}
\end{figure}

\subsection{Time-Conversion Mechanism}
Many simulated environments~\cite{zhou2024hazard} remain effectively frozen during LLM inference. This evaluation artifact can systematically overestimate the performance of slow planners in dynamic scenes.
In contrast, TCM better approximates real-time constraints by advancing the environment during deliberation. It therefore requires agents to balance decision quality against responsiveness, as extended reasoning reduces the time available for rescue operations.
Let $\mathrm{FPS}$ denote the simulator frame rate and $t_{inf}$ denote the wall-clock time, in seconds, consumed by the agent's decision-making process. The inference-frame cost \(F_{inf}\) is defined as:
\begin{equation}
F_{inf} = \lceil  t_{inf} \times \mathrm{FPS} \rceil.
\label{eq:finf}
\end{equation}

After each decision, we advance the simulator by \(F_{inf}\) frames, ensuring that the environment progresses while the agent deliberates.
Concretely, the agent acquires an observation at time $t_0$ and reasons about its next action $a_{t_0}$, incurring a decision latency of $t_{inf}$ seconds.
During the interval $[t_0,t_0+t_{inf})$, the agent remains stationary while the environment evolves autonomously. This models a blocking inference protocol in which action execution begins only after the decision is returned.
The action $a_{t_0}$ is therefore executed at $t_0+t_{inf}$, increasing the likelihood that the decision is based on stale information.

\subsection{Rapid-Reflex Async-Reflect Agent}~\label{sec:rrara}
As shown in Figure~\ref{fig:framework}, \met{} couples an immediate, low-latency reflex controller with an asynchronous high-level reflector. 
The Rapid-Reflex layer is responsible for real-time action selection.
While the low-level executor carries out the action selected by the Rapid-Reflex layer, the Async-Reflect layer monitors changes in the environment and detects whether the current plan or target has become outdated. When substantial state drift is observed, the reflector issues corrective feedback, allowing the agent to revise its target or future subgoals without incurring pause-to-think delays.

In our initial implementation, the Rapid-Reflex layer is instantiated as a deterministic rule-based policy $\pi$. At timestep $t$, the next action is selected as
\begin{equation}
a_t = \pi(O_t, C_t, M_t),
\label{eq}
\end{equation}
where $O_t$ denotes the set of currently perceived objects, including their identifiers, categories, poses, and associated hazard cues. $C_t$ encodes task-level context, including the hazard type $h \in {\mathrm{fire}, \mathrm{flood}, \mathrm{wind}}$ and the value function over object categories. $M_t$ summarizes local map metadata, including the explored mask, occupancy and height layers, and the traversability graph with precomputed A* geodesic distances.
Based on predefined selection rules and cached spatial information, the controller prioritizes nearby candidate objects and produces actions with constant-time decision latency.
Importantly, \met{} is not restricted to this specific rule-based implementation. As a general framework, any policy capable of producing reliable actions with near-constant decision latency can serve as the Rapid-Reflex backbone.

\subsection{LLM-PrePlanner}
To improve upon simple rule-based or greedy Rapid-Reflex policies, we propose the LLM-PrePlanner, which generates an ordered, object-centric sequence of high-level subgoals over a future decision horizon.
Because consecutive observations often contain highly similar contextual information, repeatedly querying the LLM introduces additional latency and redundant computation while underutilizing its long-horizon reasoning ability.
By providing multi-step foresight, the LLM-PrePlanner amortizes planning cost across the decision horizon, reducing per-step inference overhead while preserving responsiveness.

At each planning instant $t$, the pre-planner consumes a schema-formatted prompt
$(O_{t}, C_{t},M_t)$.
The input is serialized into a compact, tabular format and is paired with a deterministic output grammar enforcing that the model output consists solely of object IDs separated by commas.
Formally, the LLM-PrePlanner generates a ranked, object-oriented sequence of subgoals:
\begin{equation}
S_t = \big(p_1,p_2,\ldots,p_k\big),
\qquad
p_i =
\big(
\text{walk\_to}(o_i),~
\text{pick\_up}(o_i),~
\text{drop}(o_i)
\big),
\label{eq:llm_preplanner}
\end{equation}
where each $p_i$ represents an executable subgoal directed toward a specific object $o_i$. The top-ranked subgoal $p_1$ is dispatched to the low-level executor, which executes the corresponding atomic actions sequentially. After completing one high-level subgoal, the executor proceeds to the next cached subgoal.

\subsubsection{Deterministic Staleness Score.}
To determine whether a cached plan remains valid, we measure the state divergence between the planning snapshot at frame $\tau$ and the current state at frame $t$. Each snapshot records the frame index, agent pose, object states, and hazard field. The staleness score is defined as
\begin{equation}
\mathrm{Stale}(t,\tau)
=
\lambda_T S_T
+
\lambda_A S_A
+
\lambda_O S_O
+
\lambda_H S_H,
\label{eq:staleness_score}
\end{equation}
where $S_T$, $S_A$, $S_O$, and $S_H$ measure temporal change, agent motion, object-state change, and hazard-field variation, respectively.

For each object $o$, we define
\[
z_t(o)=
\left(
\mathrm{visible}_t(o),
\mathrm{pos}_t(o),
\mathrm{rescued}_t(o)
\right).
\]
The value-weighted object-change score is
\begin{equation}
S_O =
\frac{
\sum_{o\in\mathcal{O}_t\cup\mathcal{O}_\tau}
w(o)\mathbb{I}[z_t(o)\neq z_\tau(o)]
}{
\sum_{o\in\mathcal{O}_t\cup\mathcal{O}_\tau} w(o)
},
\qquad
w(o)=1+\beta\max(v_t(o),v_\tau(o)),
\end{equation}
where $\beta$ controls the influence of high-value objects.

When hazard maps are available, hazard variation is computed over their shared valid region $\mathcal{G}$:
\begin{equation}
S_H =
\frac{1}{|\mathcal{G}|}
\sum_{g\in\mathcal{G}}
\frac{|H_t(g)-H_\tau(g)|}
{\max(\|H_t\|_\infty,\|H_\tau\|_\infty,\epsilon)},
\end{equation}
where $\epsilon$ prevents division by zero. If no hazard map is available, we set $S_H=0$.

The cached plan is discarded and regenerated when
\begin{equation}
\mathrm{Replan}
=
\mathbb{I}
\left[
\mathrm{Stale}(t,\tau)\geq\delta
\right],
\label{eq:replan}
\end{equation}
where $\delta$ is the staleness threshold. This prevents outdated plans from guiding the agent after substantial environmental change.

\subsection{Async Reflect Layer}
A purely Rapid-Reflex controller achieves minimal decision latency but lacks deliberative reasoning. It often results in myopic selection of immediately visible, low-value targets and brittle performance under environmental change. We therefore treat it as a speed-oriented baseline rather than a competitive policy for complex, time-critical settings.
To address these limitations, we introduce the Async-Reflector as an interface between high-level reasoning and low-level execution.
While the executor acts, the LLM-based reflector continuously evaluates the current action using updated perceptual and task context, issuing targeted corrections only when necessary.

Let \(a_{t}\) denote the currently executed low-level action proposed by the Rapid-Reflex controller at time \(t\). At wall-clock \(t'\!\ge\!t\), the agent obtains the latest observation \(O_{t'}\) and task context \(C_{t'}\) (e.g., target status, agent state). 
These inputs are summarized in a structured prompt containing salient scene factors, including object hypotheses, hazard cues, proximity and visibility information, local traversability, task constraints, and target-value information encoding category-dependent utilities and value decay under fire, flood, or wind conditions.

The LLM-based reflector consumes  \((a_t,O_{t'},C_{t'})\) and returns a validation flag \(valid_{t'}\) and optionally a correction \( a_{new}\): 
\begin{equation}
(valid_{t'}, a_{new}^*) = \mathcal{V} (a_t,O_{t'},C_{t'}) \qquad valid_{t'}\in\{true,false\},
\label{eq:reflect}
\end{equation}
where $\mathcal{V}$ denotes the validation function,  $valid_{t'} $ indicates whether continuing $a_t$ remains appropriate under the most recent observation $O_{t'}$. 
If $valid_{t'}=\text{true}$, the agent continues executing $a_t$ without incurring switch latency. Otherwise, it requests a new action \(a_{new}\) as illustrated in Figure~\ref{fig:teaser}.  The star signal (\(*\)) indicates that 
\(a_{new}\) only proposed when \(valid_{t'} = false\).  The action at a new time \(t'\) is defined as:

\begin{equation}
a_{t'}=
\begin{cases}
a_t, & \text{if } valid_{t'}=1,\\
a_{new}, & \text{if } valid_{t'}=0.
\end{cases}
\label{eq:reflect_output}
\end{equation}

The validation function $\mathcal{V}$ serves as the core of the Async-Reflector. It verifies whether the incumbent action remains contextually valid given updated sensory input. In essence, $\mathcal{V}$ acts as an asynchronous critic operating in parallel with the reflex controller. It performs relational reasoning over $(a_t,O_{t'},C_{t'})$ to detect semantic or physical inconsistencies between the intended action and the current world state. These conditions include object unavailability, path obstruction, safety violations, and emerging higher-value opportunities. While the current action is being executed, the reflector may continue evaluating its validity using updated observations. This policy maintains low blocking latency by defaulting to action persistence while enabling corrections when the incumbent action becomes inappropriate.

\begin{figure}[t]
\vspace{-12pt}
\centering
\includegraphics[width=0.9\linewidth]{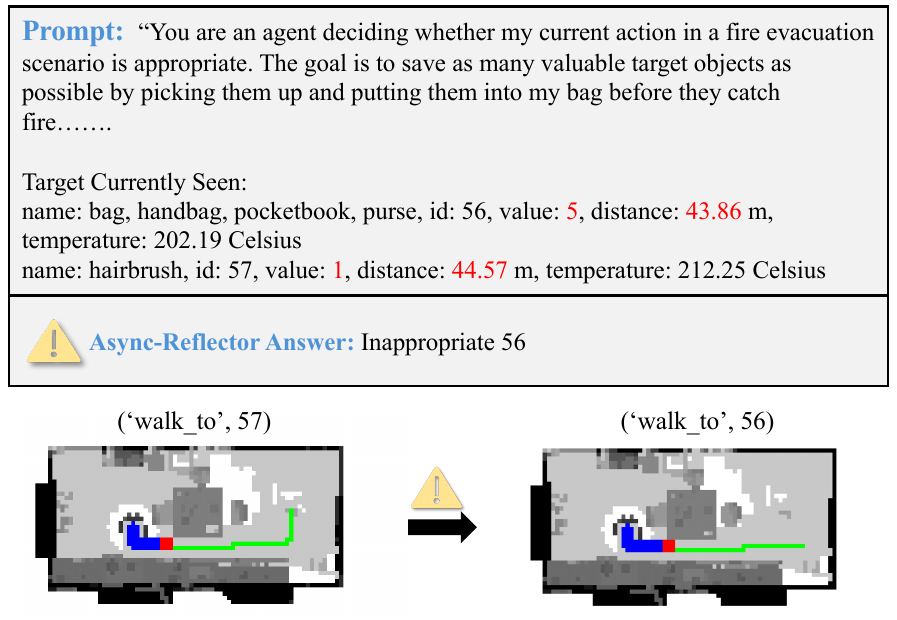}
\caption{Overview of the embodied agent architecture proposed in this work, spanning environmental perception, decision-making via the \met{} module, and low-level action execution.  }
\label{fig:result_plan}
\vspace{-12pt}
\end{figure}

\section{Experimental Setup}
\label{sec:experimental_setup}

We evaluate RRARA and all baselines on the fire, flood, and wind tasks under a unified experimental protocol, with the simulator operating at 30 FPS. Unless otherwise specified, RRARA uses a rule-based Rapid-Reflex policy and GPT-3.5 \cite{gpt3} as the Async-Reflector. Complete implementation details, including hardware, used LLM configurations \cite{llama,gpt4,gpt5,claude3,gemini2023}, semantic segmentation, supervised fine-tuning, episode limits, and hyperparameters, are provided in the supplementary material.


For fire and flood, we use four indoor rooms; for wind, we use four outdoor regions. Each room or region contains 25 scenarios with varied object placements, hazard states, and agent initial positions, resulting in 100 scenes per hazard type. We hold out one environment for testing and use the remaining three for training, yielding 75 training scenes and 25 test scenes per task.

Object perception is represented using semantic masks, and each episode specifies a set of target objects. For fire and flood, targets are sampled from 22 object categories, while wind scenarios use 11 categories. Each target object is associated with task-relevant attributes, including value, water resistance, ignition temperature, and wind susceptibility. Objects are assigned either a high initial value of \(v=5\) or a low initial value of \(v=1\). Under hazard exposure, an object's value is reduced if it is damaged and becomes zero if it is destroyed.

Formally, for each target object \(o \in \mathcal{O}\), let \(R(o) \in \{0,1\}\) indicate whether the object is successfully rescued, \(D(o) \in \{0,1\}\) indicate whether it is damaged, and \(v(o)\) denote its initial value. The environment-adjusted value is defined as
\begin{equation}
\tilde{v}(o) =
\begin{cases}
0, & \text{if \(o\) is destroyed by the hazard},\\
\frac{1}{2}v(o), & \text{if \(D(o)=1\)},\\
v(o), & \text{otherwise}.
\end{cases}
\label{eq:adjusted_value}
\end{equation}

For episode \(e\), the value rescue rate is defined as
\begin{equation}
\operatorname{VR}_e =
\frac{
    \sum_{o \in \mathcal{O}_e} R_e(o)\tilde{v}_e(o)
}{
    \sum_{o \in \mathcal{O}_e} v_e(o)
}
\in [0,1].
\label{eq:value_rescue_rate}
\end{equation}

The average value rescue rate across \(N\) episodes is
\begin{equation}
\operatorname{VR} =
\frac{1}{N}
\sum_{e=1}^{N}
\operatorname{VR}_e.
\label{eq:average_value_rescue_rate}
\end{equation}

We measure the proportion of time lost to inference using the inference latency ratio:
\begin{equation}
\operatorname{ILR} =
\frac{
    t_{\mathrm{inf}}
}{
    t_{\mathrm{inf}} + t_{\mathrm{action}}
}.
\label{eq:inference_latency_ratio}
\end{equation}
Here, \(t_{\mathrm{inf}}\) denotes the wall-clock inference or decision latency that delays action execution, and \(t_{\mathrm{action}}\) denotes the duration of the corresponding physical or low-level action execution. For asynchronous agents, inference performed concurrently with action execution is not included in \(t_{\mathrm{inf}}\).

\section{Results}

\begin{table}[t]
\centering
\caption{Impact of TCM on average value rate. ``TCM'' denotes evaluation with inference latency, and $\Delta=\mathrm{TCM}-\mathrm{Base}$. The average delta is computed across the fire, flood, and wind environments.}
\label{tab:overall_results}

\begin{adjustbox}{max width=1\textwidth}
\begin{tabular}{
l
S[table-format=1.2] S[table-format=1.2] c
S[table-format=1.2] S[table-format=1.2] c
S[table-format=1.2] S[table-format=1.2] c
c
}
\toprule
\multirow{2}{*}{Agent}
& \multicolumn{3}{c}{\textbf{Fire}}
& \multicolumn{3}{c}{\textbf{Flood}}
& \multicolumn{3}{c}{\textbf{Wind}}
& \multirow{2}{*}{\makecell{\textbf{Avg.}\\$\boldsymbol{\Delta}$}} \\
\cmidrule(lr){2-4}
\cmidrule(lr){5-7}
\cmidrule(lr){8-10}

& {Base} & {TCM} & {$\Delta$}
& {Base} & {TCM} & {$\Delta$}
& {Base} & {TCM} & {$\Delta$}
& \\
\midrule

\multicolumn{11}{l}{\emph{Classic agents}} \\

MCTS
& 0.75 & 0.25 & $-0.50$
& 0.40 & 0.15 & $-0.25$
& 0.15 & 0.02 & $-0.13$
& -0.29 \\

Rule
& 0.33 & 0.35 & $+0.02$
& 0.25 & 0.23 & $-0.02$
& 0.00 & 0.00 & {--}
& 0.00 \\

Greedy
& 0.08 & 0.11 & $+0.03$
& 0.08 & 0.09 & $+0.01$
& 0.00 & 0.00 & {--}
& $+0.01$ \\

\midrule
\multicolumn{11}{l}{\emph{LLM agents}} \\

GPT-3.5
& 0.65 & 0.44 & $-0.21$
& 0.36 & 0.24 & $-0.12$
& 0.09 & 0.06 & $-0.03$
&  $-0.12$ \\

GPT-4
& 0.71 & 0.22 & $-0.49$
& 0.44 & 0.21 & $-0.23$
& 0.20 & 0.04 & $-0.16$
& $-0.29$ \\

GPT-5
& 0.73 & 0.35 & $-0.38$
& 0.47 & 0.22 & $-0.25$
& 0.24 & 0.07 & $-0.17$
&  $-0.27$ \\

Llama-2-7b-hf
& 0.60 & 0.05 & $-0.55$
& 0.31 & 0.05 & $-0.26$
& 0.04 & 0.00 & $-0.04$
& $-0.28$ \\

Claude-3.7-sonnet
& 0.77 & 0.10 & $-0.67$
& 0.48 & 0.10 & $-0.38$
& 0.21 & 0.02 & $-0.19$
& $-0.41$ \\

Claude-4.6-sonnet
& 0.71 & 0.18 & $-0.53$
& 0.43 & 0.11 & $-0.32$
& 0.24 & 0.02 & $-0.22$
& $-0.36$ \\

Gemini-3.5-flash
& 0.69 & 0.22 & $-0.47$
& 0.40 & 0.15 & $-0.25$
& 0.09 & 0.03 & $-0.06$
&  $-0.26$ \\

\bottomrule
\end{tabular}
\end{adjustbox}
\end{table}


\begin{wrapfigure}{r}{0.6\textwidth}
    \vspace{-15pt}
    \centering
    \includegraphics[width=\linewidth]{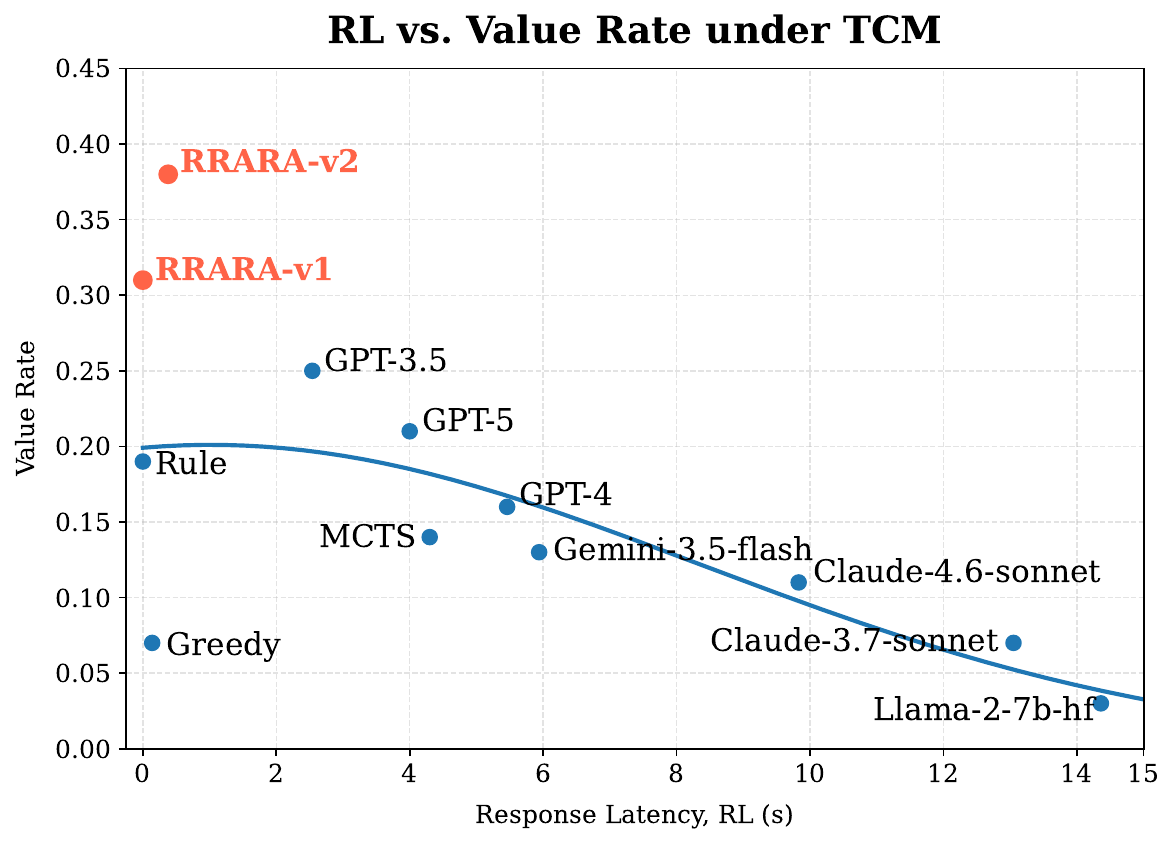}
    \caption{
    Relationship between response latency and value rate under TCM.
    }
    \vspace{-20pt}
    \label{fig:rl-value-rate}
    
\end{wrapfigure}

Under the latency-aware TCM setting, the performance ranking changes substantially. Classical finite-state agents remain largely stable because their decisions are nearly instantaneous.
In contrast, methods that perform strongly in the base setting degrade once inference time is converted into control delay. By advancing the simulator during inference, TCM reduces the number of available actions, allows hazards to evolve, and increases plan staleness.
Among the evaluated LLM agents, GPT-3.5 exhibits the smallest average reduction of 0.12 in value rate, whereas Claude-3.7-Sonnet exhibits the largest average reduction of 0.41.
We use a rule-based agent as the Rapid-Reflex backbone in RRARA-v1 and the LLM-PrePlanner in RRARA-v2. Both variants employ a fine-tuned GPT-3.5 reflector, balancing reasoning quality with favorable throughput under TCM. As shown in Table~\ref{tab:latency-budget}, both \met{} variants substantially outperform the baselines.

\begin{table}[t]
\centering
\caption{ Response Latency, Latency-to-Action Ratio, Average number of actions and value rate across various agents under TCM.
}
\label{tab:latency-budget}
\begin{adjustbox}{max width=1.1\textwidth}
\begin{tabular}{lcccc}
\toprule
\textbf{Agent}
& \textbf{RL (s) $\downarrow$}
& \textbf{Avg. Actions $\uparrow$}
& \textbf{LAR (\%) $\downarrow$}
& \textbf{Value Rate $\uparrow$} \\
\midrule
Rule
& 0.00 & 26 & 0  & 0.19 \\
Greedy
& 0.14 & 20 & 6  & 0.07 \\
GPT-3.5
& 2.54 & 11 & 52 & 0.25 \\
GPT-5
& 4.00 & 9  & 67 & 0.21 \\
MCTS
& 4.30 & 8  & 61 & 0.14 \\
GPT-4
& 5.46 & 7  & 72 & 0.16 \\
Gemini-3.5-flash
& 5.94 & 6 & 73 & 0.13 \\
Claude-4.6-sonnet
& 9.83 & 5 & 86 & 0.11 \\
Claude-3.7-sonnet
& 13.05 & 4 & 89 & 0.07 \\
Llama-2-7b-hf
& 14.36 & 4 & 89 & 0.03 \\
\midrule
LLM-PrePlanner
& 0.40 & 20 & 17 & 0.40 \\
\met{}-v1
& 0.00 & 27 & 0 & 0.37 \\ 
\met{}-v2
& 0.38 & 25 & 17 & 0.44 \\ 
\bottomrule
\end{tabular}
\end{adjustbox}
\end{table}

For asynchronous agents, the reported Response Latency represents the blocking delay before the executor can begin or continue acting. It does not include the total computation time of the background reflector, which runs concurrently with action execution.
RRARA-v1 achieves an average value rate of 0.37, exceeding the strongest baseline. RRARA-v2 achieves the strongest overall average performance, reaching a value rate of 0.44. The difference mainly arises from subgoal quality: the rule-based backbone provides high throughput but remains myopic, whereas the LLM-PrePlanner maintains low blocking latency while generating more effective subgoals. The asynchronous reflector adjusts actions when the current decision becomes inconsistent with updated environmental observations. The LLM-PrePlanner also outperforms \met{}-v1, indicating that explicit long-horizon scheduling benefits extended tasks such as HAZARD. Finally, coupling the LLM-PrePlanner with the Async-Reflector in \method{} yields a value rate of 0.44, the highest overall result.
This combination integrates strategic object ordering with asynchronous validation, enabling timely corrections in latency-sensitive HAZARD scenarios.

\subsection{Expert-Guided Fine-Tuning}

We fine-tune GPT-3.5 across the fire, flood, and wind scenarios using GPT-5 as the expert model, the results are shown in Table~\ref{tab:GPT35-finetune-avg}. Motivated by GPT-5's strong zero-shot performance, we use it to generate planning trajectories under two data regimes: 30 selected expert demonstrations and 1{,}000 expert demonstrations.
With 30 training examples, GPT-3.5 improves its average value rate from 0.25 to 0.30. Scaling to 1{,}000 examples further increases the value rate to 0.35, yielding an improvement of 0.10 over the pretrained model. Fine-tuning also slightly reduces inference latency, preserving efficiency under TCM.

\begin{table}[t]
\centering
\caption{GPT-3.5 fine-tuning results averaged across the fire, flood, and wind tasks.}
\label{tab:GPT35-finetune-avg}
\setlength{\tabcolsep}{6pt}
\begin{tabular}{lc}
\toprule
\textbf{Fine-tune}  & \textbf{Avg.\ Value Rate} \\
\midrule
Pretrained                  & 0.25 \\
30 training examples        & 0.30 \\
1{,}000 training examples   & 0.35 \\
\bottomrule
\end{tabular}
\end{table}

\paragraph{Limitations.}
TCM models blocking inference by advancing the environment while keeping the agent stationary, whereas alternative execution protocols may allow an agent to continue its previous action during deliberation. For asynchronous agents, the reported Response Latency measures only action-blocking delay and does not capture the total background computation performed by the planner and reflector. Moreover, some experiments rely on remotely hosted LLMs, whose response times may vary with network conditions, server-side load, and serving infrastructure, limiting the reproducibility of absolute latency measurements. Future work should evaluate alternative execution protocols, controlled local-inference settings, and deployment on physical robots.

\section{Conclusion}
This work identifies inference latency as a key limitation of LLM-based embodied agents in dynamic, high-risk environments. We introduce the Time Conversion Mechanism (TCM), Response Latency, and Latency-to-Action Ratio to quantify its impact on real-time performance. Results show that stronger reasoning can be offset by longer response delays.
To address this trade-off, we propose \method{} (\met{}), which combines low-latency reflexive control with asynchronous reflection. The LLM-PrePlanner and expert-guided fine-tuning further improve decision quality while preserving responsiveness.
Overall, our work provides a unified latency-aware evaluation framework and a practical architecture for reliable embodied decision-making under real-world temporal constraints.

\bibliographystyle{splncs04}
\bibliography{ref}

\input{supp}
\end{document}

%% file: supp.tex
\clearpage
\appendix
\begin{center}
\Large\bfseries Reflex First, Reflect Later: Latency-Aware Embodied LLM Agents for Dynamic Response \\
\vspace{2mm}
\large Appendix
\vspace{1em}
\end{center}

\section{Implementation Details}
Experiments are conducted on a system equipped with an Intel Core i7-11700 CPU and a single NVIDIA GeForce RTX 3090 GPU. We evaluate all HAZARD tasks, including fire, flood, and wind, with the simulator running at 30 FPS. Unless otherwise specified, RRARA uses a rule-based Rapid-Reflex policy and GPT-3.5 as the Async-Reflector.

We evaluate LLM-based agents using Llama-2-7B-HF~\cite{llama}, GPT-family models~\cite{gpt3,gpt4,gpt5}, Claude~\cite{claude3}, and Gemini 3.5 Flash~\cite{gemini2023}. Inference uses a maximum of 512 output tokens, a temperature of 0.7, and top-$p=1.0$. For GPT-5, verbosity is set to medium and reasoning effort to minimal.

For semantic segmentation, we fine-tune Mask R-CNN~\cite{rcnn} using the OpenMMLab detection framework. For each hazard task, 200 images with ground-truth segmentation masks are collected for training.

LLM-PrePlanner and Reflector variants requiring supervised adaptation use GPT-3.5-Turbo-0125 fine-tuned on 1{,}000 expert demonstrations. Training is performed for two epochs with a batch size of 1 and a learning-rate multiplier of 2. Episodes are capped at 1{,}500 frames for fire and flood and 3{,}000 frames for wind. HAZARD baselines are reproduced using their default hyperparameters.
For the LLM-PrePlanner, the staleness threshold is set to $\delta=0.6$, triggering replanning only after substantial environmental change. This reduces unnecessary LLM calls while ensuring that outdated plans are updated under significant state drift.

\section{Qualitative Analysis}

\begin{figure}
\vspace{-5pt}
\centering
\includegraphics[width=1.1\textwidth]{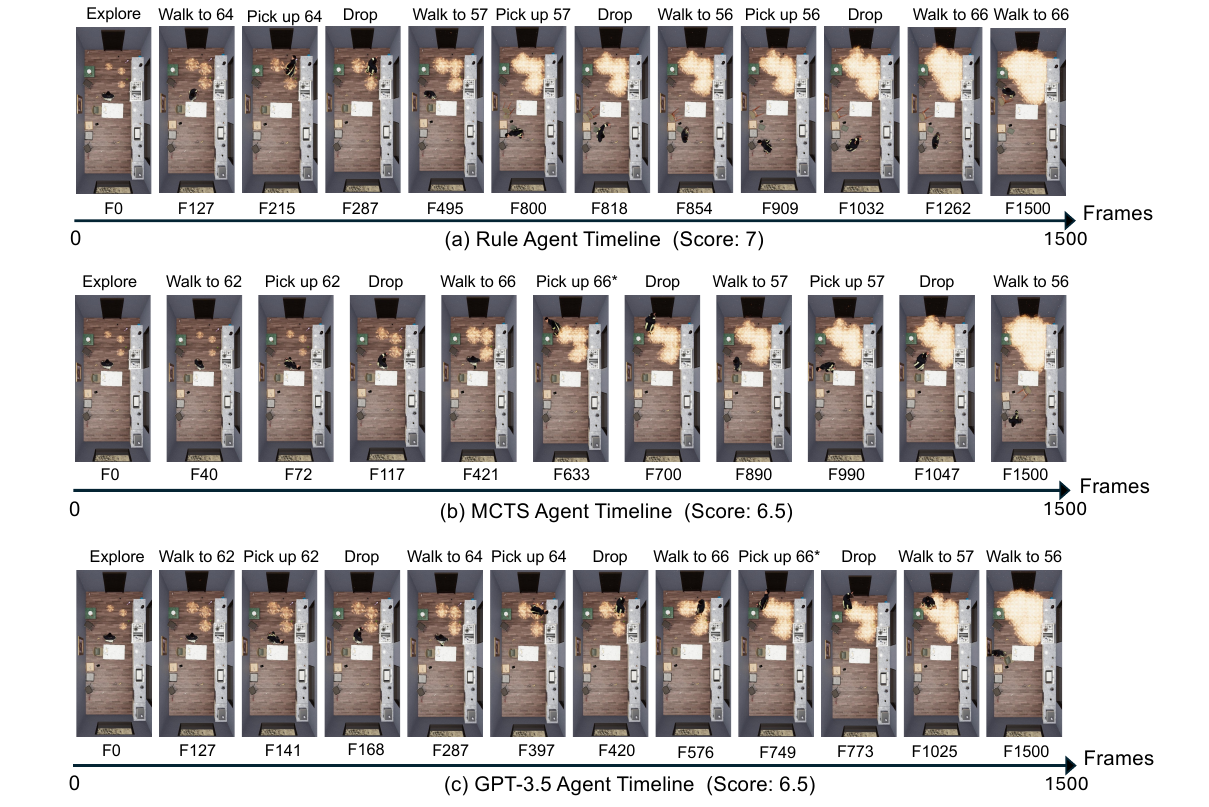}
\caption{Visualization of the frame timeline showing how different agents, including rule, MCTS, and GPT-3.5, rescue objects in the fire scenario.
  Asterisks (*) indicate damaged objects. The score represent the total value object that has been rescued.  The total object value in the scene is 13, where Objects 62 and 56 are high-value (5 each), and Objects 66, 57, and 64 are low-value (1 each).``F1'' denotes the first decision frame, and the corresponding action taken by the agent is shown below each frame.}
\label{fig:v1}
\end{figure}

\begin{figure}
\vspace{-10pt}
\centering
\includegraphics[width=1.1\textwidth]{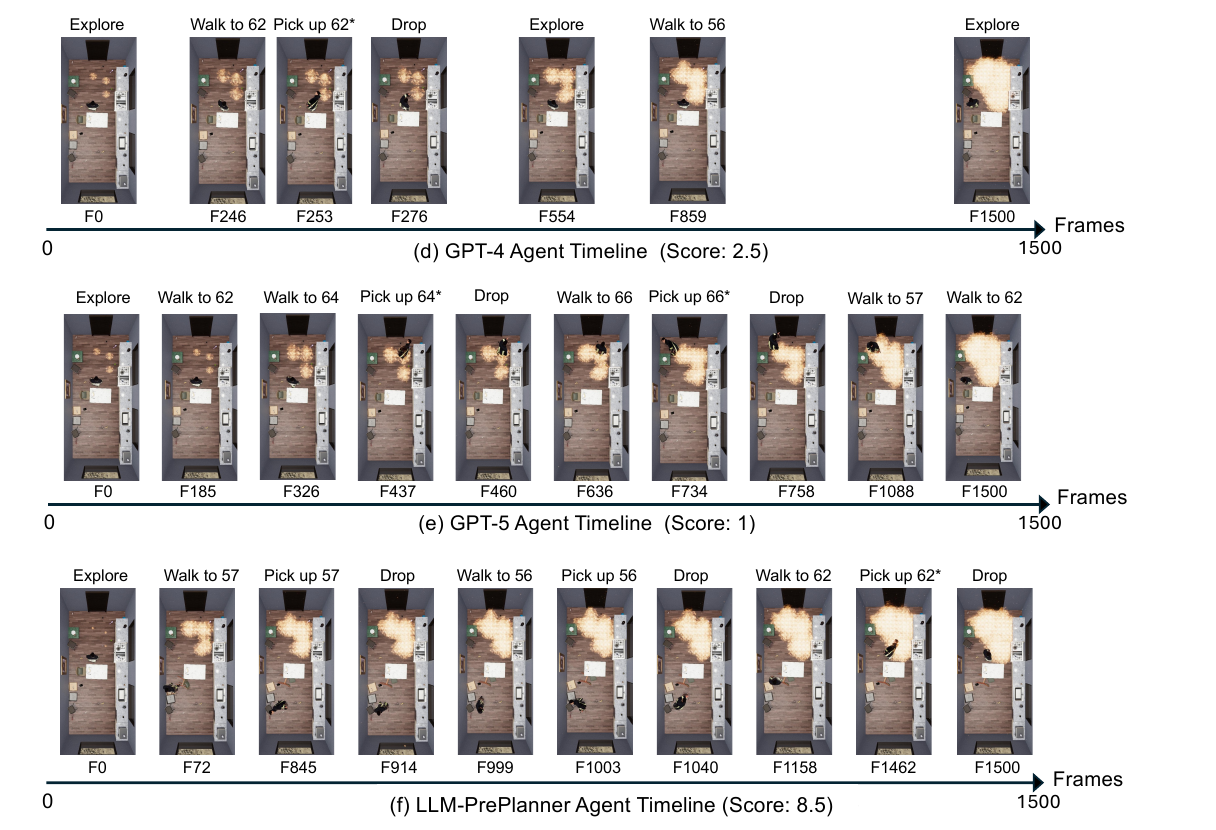}
\caption{Visualization of the frame timeline showing LLM-based agents (GPT-4, GPT-5) and the proposed LLM-PrePlanner rescue objects in the fire scenario. }
\label{fig:v2}
\end{figure}

As shown in Figure~\ref{fig:v1} and \ref{fig:v2}, the agent performs the rescue task under the fire scenario.
Although baseline agents manage to collect some objects, their actions are often inefficient due to response latency or failure to account for object value. 
Because each object carries a different score, collecting items as early as possible does not necessarily produce the best outcome.

GPT-3.5 tends to prioritize objects that near the danger.
While such behavior may be advantageous in less time-critical scenarios, it underperforms when strict temporal constraints are present, and its neglect of object value explains its moderate result of 6.5, comparable to MCTS.
GPT-4 and GPT-5 spend much more time reasoning before acting, which causes most objects to become damaged and reduces the number of actions they can take.

A recurring challenge observed in LLM-based agents is nondeterminism, which leads to frequent plan revisions during execution.
Such instability causes the agent to repeatedly switch its target objects, wasting substantial time in indecisive transitions and ultimately resulting in no successful object collection. 
As shown in Figure~\ref{fig:v2} (e), the GPT-5 agent continuously alters its intended goal, losing time as a trade-off between planning depth and timely execution.

In contrast, the LLM-PrePlanner mitigates this issue by generating a rescue schedule in advance. By following a fixed plan, it maintains consistent behavior and prioritizes rescuing high-value objects first. 
Although it collects fewer items overall, it achieves a higher total value of 8.5. However, nearby objects may be neglected due to the rigidity of the pre-defined plan.
Finally, our proposed \met{} framework balances efficiency with adaptability. The reflective module provides real-time feedback, prompting the agent to pick up a nearby Object 66 before moving to the opposite side of the room, thereby optimizing the action order. This result demonstrates that \met{} executes actions efficiently while preserving reasoning ability to prefer closer or higher-value objects. Consequently, \met{} achieves the best overall performance in this scenario.

\begin{figure}
\centering
\includegraphics[width=0.95\textwidth]{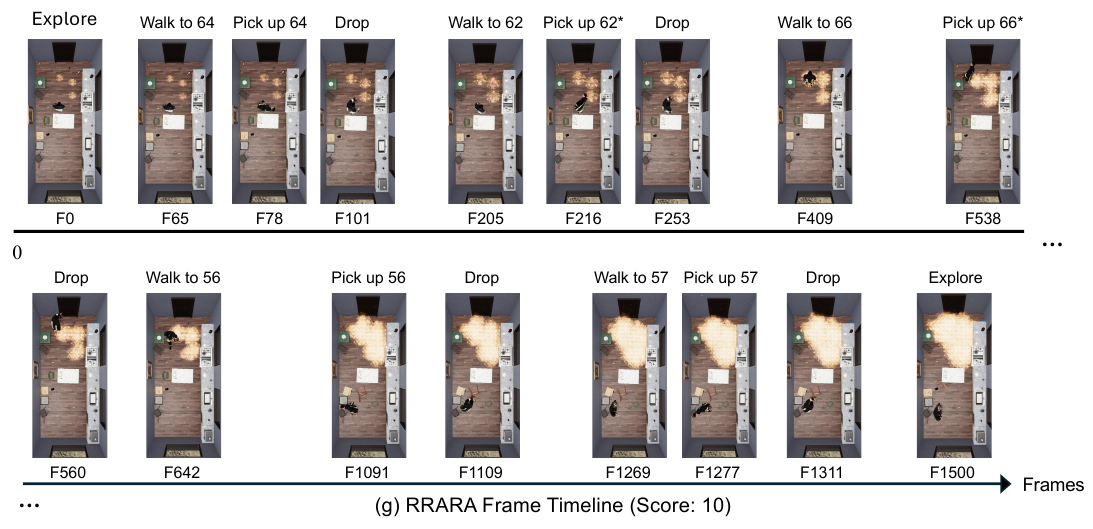}
\caption{Visualization of the frame timeline showing the RRARA agent to rescue object in the fire scenario.}
\label{fig:v3}
\end{figure}